\documentclass[10pt,twocolumn,letterpaper]{article}

 \usepackage{cvpr}              

\usepackage{cuted} 

\usepackage{currfile} 

\usepackage{caption} 
\definecolor{cvprblue}{rgb}{0.21,0.49,0.74}
\usepackage[pagebackref,breaklinks,colorlinks,allcolors=cvprblue]{hyperref}

\def\paperID{15061} 
\def\confName{CVPR}
\def\confYear{2027}
 
\newcommand{\systemname}{{KROMA}~}

\usepackage{booktabs}
\usepackage{multirow}
\usepackage[table]{xcolor}
\usepackage{graphicx}

\title{KROMA: \underline{K}nowledge-\underline{R}etaining \underline{O}nline \underline{M}ultimodal \underline{A}daptation for 3D Object Detection in Changing Driving Conditions}

\author{
Ye Tian$^{1}$ \quad
Md Mizanur Rahaman Nayan$^{2}$ \quad
Quanling Zhao$^{1}$ \quad
Tianqi Zhang$^{1}$ \\
Flavio Ponzina$^{3}$ \quad
Azad Naeemi$^{2}$ \quad
Tajana Rosing$^{1}$ \\[3pt]
$^{1}$University of California San Diego, La Jolla, CA, USA \\
$^{2}$Georgia Institute of Technology, Atlanta, GA, USA \\
$^{3}$San Diego State University, San Diego, CA, USA \\[2pt]
{\tt\small \{yet002,quzhao,tiz014,tajana\}@ucsd.edu} \\
{\tt\small \{mizanur.rahaman,azad\}@gatech.edu} \quad
{\tt\small fponzina@sdsu.edu}
}

\begin{document}
\maketitle

\begin{abstract}
Autonomous driving requires reliable multimodal 3D object detection under changing environmental conditions.
Although online adaptation enables detectors to learn from incoming observations, existing methods struggle to identify and reuse relevant historical knowledge while preserving previously acquired capabilities and retaining useful adaptations.
We propose KROMA, a supervised online adaptation framework integrating context-aware knowledge reuse with knowledge-preserving learning. KROMA organizes detector states according to compositional semantic attributes, enabling relevant historical knowledge to be retrieved and reused as conditions evolve.
It reactivates retained states for recurring contexts, while selecting suitable historical states for new contexts based on current detection performance.
Context-specific residual adaptation and knowledge-preserving mechanisms mitigate knowledge interference, while selective state retention preserves useful updates for future reuse.
Extensive experiments demonstrate that KROMA achieves $\mathrm{AP}_{\mathrm{3D}}^{0.3}$ of 73.55\%, outperforming the strongest baselines by 7.7 percentage points, while achieving a favorable accuracy--efficiency trade-off.
Further analysis demonstrates its advantages under challenging driving conditions, including heavy snow, foggy mountain roads, and nighttime driving.
Our code will be made publicly available.
\end{abstract}

\section{Introduction}
\label{sec:introduction}

Autonomous driving relies on accurate 3D perception to localize surrounding objects and support driving decisions under changing conditions.
Recent studies have demonstrated the advantages of fusing complementary camera, LiDAR, and 4D radar observations for robust 3D object detection~\cite{paek2026availability,park2026raf}.
However, environmental changes can alter sensor reliability and data distributions, degrading detector performance~\cite{bijelic2020seeing,dong2023benchmarking}, as these changes may not be adequately represented in offline training data~\cite{eskandar2024empirical,hegde2025multimodal}.
Repeated training and redeployment are time-consuming and computationally expensive.
This motivates continual online adaptation, enabling multimodal detectors to continuously learn from new observations and update their models.

Existing online adaptation methods address distribution shifts in autonomous driving through selective updates and dynamic tuning~\cite{yoo2024and,park2025hybrid,cao2026continual}, domain-adaptive expert routing~\cite{lee2024becotta}, and historical knowledge reuse~\cite{chen2025mos,yang2025codemerge}.
However, existing methods primarily focus on single or limited sensor modalities, with an emphasis on weather changes and domain shifts.
Continual adaptation of camera--LiDAR--4D radar fusion detectors, particularly under more complex combinations of environmental changes, remains largely unexplored.

Motivated by this gap, we conduct preliminary experiments with representative online adaptation methods on K-Radar, a large-scale real-world multimodal dataset covering complex variations in weather, road types, and lighting conditions~\cite{paek2022k}.
The preliminary experiments reveal two key insights.
First, although these methods are designed for unsupervised adaptation, unsupervised updates can even degrade model performance under such diverse conditions.
The results show that substantial distribution shifts can compromise the quality of pseudo-labels generated by the source model, leading to erroneous updates and accumulated model drift, consistent with some previous findings~\cite{chen2024dpo,park2025hybrid,cao2026continual}.
Second, even with supervised data, prolonged adaptation to changing driving conditions remains vulnerable to performance fluctuations and degradation, particularly in contexts learned earlier.
This indicates that online adaptation requires a better balance between acquiring new knowledge and preserving existing capabilities.
In this work, we focus on addressing the second issue: 
How to continually acquire new knowledge while effectively preserving and reusing previously learned knowledge to maintain stable adaptation across changing environments?

Achieving this goal requires addressing two fundamental challenges.
First, reliably identifying the current driving context and selecting relevant historical knowledge are challenging.
Previously encountered conditions may reappear, while new conditions may share characteristics with earlier ones, making historical knowledge valuable for subsequent adaptation.
However, driving conditions arise from diverse combinations of environmental factors, with unknown transition boundaries and potentially abrupt changes.
This makes it difficult to distinguish contexts and select the most relevant historical knowledge.
Second, acquiring new knowledge without compromising previously learned capabilities is challenging.
Continuous updates to shared parameters can cause knowledge interference across contexts, leading to model drift and forgetting.
Moreover, not all updated model states are equally reliable or valuable for future adaptation.
Stable adaptation therefore requires protecting existing knowledge while selectively retaining useful new knowledge for future reuse.

Existing online adaptation methods provide mechanisms for model updating and historical knowledge reuse, but these mechanisms alone do not ensure that past knowledge is well matched to the current driving context or that useful adaptations are reliably retained~\cite{chen2025mos,yang2025codemerge}.
Our key insight is to organize and retain detector states according to their driving contexts, enabling relevant historical knowledge to be retrieved and reused as conditions evolve.
Guided by this insight, we propose KROMA, a framework for supervised online adaptation of multimodal 3D detectors.
We design a context-aware knowledge reuse strategy that represents driving conditions through compositional semantic attributes, including weather, road type, and lighting, and associates recognized contexts with retained detector states in semantic memory.
For previously encountered contexts, KROMA reactivates their retained states, while for new contexts, it selects a suitable historical initialization based on detection performance on current observations.
Building on the retrieved historical state, KROMA employs context-specific residual adaptation based on parameter superposition~\cite{cheung2019superposition} to limit knowledge interference.
Shared detector parameters remain frozen, while independently maintained residuals allow the active context to be updated without directly changing other retained states.
To preserve inherited knowledge during adaptation, KROMA combines regularization and context-specific rehearsal to mitigate model drift and forgetting of earlier experience.
Finally, selective state retention separates the continuously updated learner from a performance-selected accepted state, retaining the latter in semantic context memory for future reuse.

Extensive experiments demonstrate that KROMA achieves the best $\mathrm{AP}_{\mathrm{BEV}}$ and $\mathrm{AP}_{\mathrm{3D}}$ at both IoU thresholds, with $\mathrm{AP}_{\mathrm{3D}}^{0.3}$ and $\mathrm{AP}_{\mathrm{3D}}^{0.5}$ of 73.55\% and 48.62\%, outperforming the strongest baselines by 7.67 and 9.92 percentage points (pp), respectively.
KROMA also achieves 2.65 effective FPS, a 21.6\% throughput gain over Hybrid-TTA~\cite{park2025hybrid}, the strongest baseline in detection accuracy.
Across different sequences of changing driving contexts, KROMA exhibits the lowest mean absolute AP$_{\mathrm{3D}}$ deviation, demonstrating greater adaptation stability.
Furthermore, KROMA also shows substantial advantages under challenging driving conditions, particularly heavy snow, foggy mountain roads, and nighttime driving. Our main contributions are summarized as follows:
\begin{itemize}
    \item We introduce a context-aware knowledge reuse strategy to identify relevant knowledge across driving conditions.
   
    \item We design a knowledge-preserving online adaptation method that preserves detection capabilities from previous contexts while acquiring knowledge from new ones.
    
    \item Experiments across diverse driving contexts demonstrate that KROMA improves detection accuracy while achieving a favorable accuracy--efficiency trade-off.
\end{itemize}

\section{Related Work}
\label{sec:Relatedwork}

\subsection{Perception under Diverse Driving Conditions}
Multimodal 3D object detection leverages complementary information from cameras, LiDAR, and radar, yet variations in environment and sensor reliability can degrade perception performance~\cite{dong2023benchmarking,cho2026dsertroll}.
To improve robustness, existing studies have explored adaptive sensor fusion under adverse weather~\cite{bijelic2020seeing,palladin2024samfusion}, enhanced LiDAR--radar integration~\cite{huang2025l4dr,chae2025doppler}, and reliability-aware multimodal fusion~\cite{park2026raf,paek2026availability}.
Beyond robust sensor fusion, condition-aware approaches leverage environmental cues to guide feature gating and modality weighting~\cite{chae2024towards,wang2026swg}, allocate domain-specific adaptation capacity~\cite{yang2025da3d}, or select weather-specific experts and fusion branches~\cite{lin2026aw,li2026weather}.
Although these approaches improve perception robustness across diverse driving conditions, they largely rely on offline training with extensive data collected across different environments.
This reliance limits their ability to continually adapt to new or evolving driving conditions after deployment.

\begin{figure*}[t]
    \centering
    \includegraphics[
        width=\textwidth
    ]{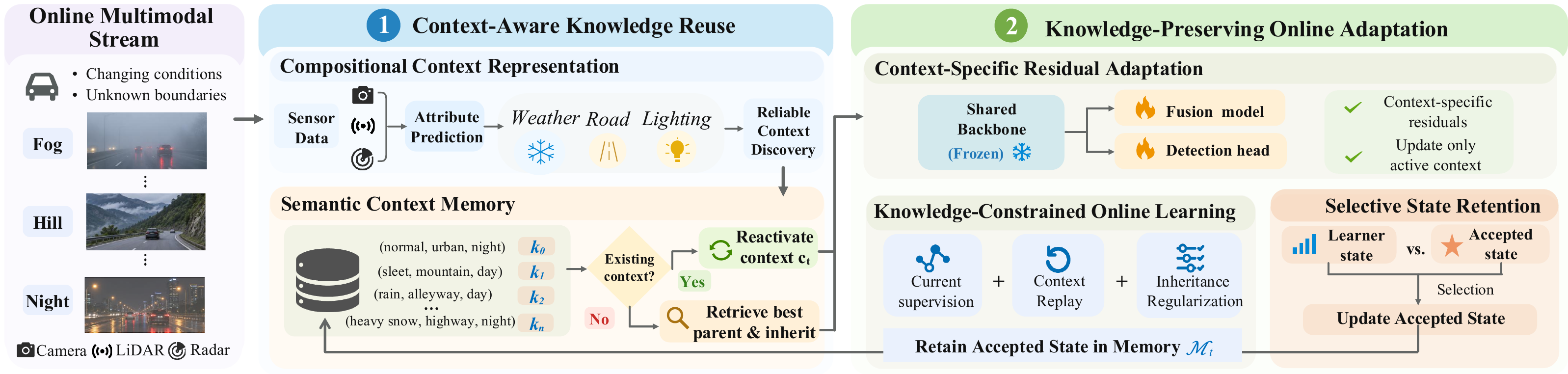}
    \caption{
    \textbf{Overview of KROMA.}
    KROMA identifies driving contexts to reactivate or retrieve historical detector states, performs knowledge-preserving online adaptation through context-specific updates, and selectively retains reliable states for future reuse.
}
    \label{fig:kroma_overview}
\end{figure*}

\subsection{Continual Adaptation and Knowledge Reuse}

To address post-deployment distribution shifts, recent studies have explored continual adaptation for 3D perception, including reliability- and uncertainty-guided adaptation for monocular detection~\cite{lin2024monotta,hu2025adaptive}, frequency-based feature disentanglement for LiDAR detection~\cite{li2026find}, and domain-aware adaptation for continual LiDAR segmentation~\cite{zhao2025d}.
Online 3D detection correction through visual prompting has also been investigated~\cite{zhang2025test}.
To handle evolving domains, existing methods mitigate adaptation instability and forgetting through parameter restoration, selective optimization, and dynamic update strategies~\cite{wang2022continual,cao2023multi,yoo2024and,chen2024dpo,yuan2024reg,cao2026continual,park2025hybrid}.
Recent approaches further explore sensitivity-guided pruning, domain-invariant training, discrepancy-aware adaptation, and environment-conditioned parameter generation~\cite{wang2025efficient,fan2025robust,song2026cd,li2025continual}.
Beyond controlling individual updates, historical knowledge reuse has gained increasing attention.
MOS~\cite{chen2025mos} and CodeMerge~\cite{yang2025codemerge} leverage historical checkpoints through model selection and composition.
BECoTTA~\cite{lee2024becotta} employs domain-adaptive expert routing, while DPCore~\cite{zhang2025dpcore} and ReservoirTTA~\cite{vray2025reservoirtta} maintain specialized knowledge for evolving and recurring domains.
Despite these advances, effectively integrating context-aware historical knowledge reuse with knowledge preservation remains challenging for multimodal 3D detection under changing driving conditions.


\section{\systemname}
\label{sec:KROMA}

\subsection{Problem Formulation and Overview}
\label{sec:overview}

We consider supervised continual adaptation of a source-pretrained multimodal 3D object detector under changing driving conditions.
The detector receives a sequential target stream $\mathcal{D}=\{(x_t,y_t)\}_{t=1}^{T}$, where $x_t$ comprises camera, LiDAR, and radar observations, and $y_t$ denotes the corresponding 3D detection annotations available for online adaptation and model selection.
Driving conditions evolve over time with unknown transition boundaries, and previously encountered conditions may recur.
Our objective is to continuously adapt to new conditions while preserving detection capabilities acquired in earlier environments.

We propose KROMA, which integrates context-aware knowledge reuse with knowledge-preserving online adaptation, as illustrated in Fig.~\ref{fig:kroma_overview}.
First, we design a context-aware knowledge reuse strategy (Sec.~\ref{sec:knowledge_reuse}) that organizes historical detector states using compositional semantic attributes, enabling direct reactivation of retained states for recurring contexts and detection-guided retrieval of suitable historical states for new contexts.
Building on the retrieved knowledge, we develop a knowledge-preserving online adaptation approach (Sec.~\ref{sec:knowledge_preserving}) that reduces interference through context-specific residual updates and knowledge-constrained learning.
Finally, we selectively retain well-performing model states to enable the reuse of relevant knowledge in future adaptation.

\subsection{Context-Aware Knowledge Reuse}
\label{sec:knowledge_reuse}

\noindent\textbf{Compositional Context Representation.}
Driving conditions arise from combinations of interacting environmental factors, and modeling every possible combination as a separate category may require extensive labeled data and costly training.
To exploit commonalities across different driving conditions, KROMA represents each context through three semantic attributes: weather, road type, and lighting.
This compositional representation captures environmental attributes shared across contexts while preserving a distinct semantic identity for each combination.

To learn and predict these attributes, we design a lightweight multimodal predictor using features from frozen camera, LiDAR, and radar encoders.
Compact modality descriptors are extracted by concatenating the channel-wise spatial means and standard deviations of their feature maps.
Since different attributes are characterized by distinct sensory cues, we construct attribute-specific representations and prediction branches.
Weather prediction combines visual appearance with LiDAR and radar features through learned modality weighting, while road-type prediction integrates visual semantics and LiDAR geometry.
Lighting prediction combines camera features with illumination statistics, including brightness, luminance variation, and color information.
Each branch produces an attribute-specific representation $\mathbf{z}_t^{(a)}$ for attribute $a$.

The predictor is pretrained with attribute-wise supervision and remains frozen during online adaptation to maintain a consistent semantic mapping.
Let $\mathcal{A}=\{w,r,l\}$ denote the attribute types and $\mathcal{Y}_a$ the label space of attribute $a$.
Each prediction head $g_a$ produces $\mathbf{p}_t^{(a)}=g_a(\mathbf{z}_t^{(a)})$.
The predicted attributes form a candidate semantic key:
\begin{equation}
    \hat{v}_t^{(a)}
    =
    \arg\max_{v\in\mathcal{Y}_a}
    p_t^{(a)}(v),
    \qquad
    \hat{k}_t =
    \big(
    \hat{v}_t^{(w)},
    \hat{v}_t^{(r)},
    \hat{v}_t^{(l)}
    \big).
    \label{eq:semantic_context}
\end{equation}
The semantic key $\hat{k}_t$ provides a structured index for organizing historical model states by driving condition, supporting the retrieval and reuse of relevant knowledge when these conditions recur.

\noindent\textbf{Reliable Context Discovery.}
The candidate semantic key $\hat{k}_t$ describes the current driving conditions, but transient prediction errors can activate unrelated model states and direct subsequent updates to the wrong context.
To distinguish persistent context changes from temporary fluctuations, KROMA combines prediction confidence with temporal consistency.
Within a sliding observation window, each frame votes for its complete semantic key only when all attribute predictions exceed a confidence threshold.
A context transition is confirmed only when the same key receives sufficient votes in consecutive windows, preventing transient prediction fluctuations from triggering unnecessary model switches.

To organize historical knowledge, KROMA maintains a semantic context memory $\mathcal{M}_{t-1}$ that associates established semantic keys with their corresponding model states.
Given a confirmed key $k_t$, a matching memory entry allows the previously encountered context to be reactivated.
Otherwise, KROMA registers a new context whose model state will be initialized using relevant historical knowledge.
This distinction enables direct reuse when driving conditions recur while accommodating new combinations of environmental attributes.
During context transitions, observations may arrive before their semantic identity can be reliably established.
Updating before the context is confirmed may lead to inappropriate updates to the model, while discarding these observations would waste potentially useful adaptation data.
Thus, during uncertain context transitions, KROMA retains the active context for inference, suspends parameter updates, and stores high-confidence observations in a bounded transition buffer.
Once the context is confirmed, buffered observations matching $k_t$ form $\mathcal{B}_t$, from which detection-eligible samples are allocated to online adaptation and context-specific model selection.
In this way, KROMA preserves observations collected during context transitions while reducing the risk of uncertain data interfering with acquired knowledge.

\noindent\textbf{Historical Knowledge Retrieval.}
When a new context is identified, selecting an appropriate historical state is important for stable adaptation.
Restarting from the source detector overlooks previously acquired knowledge, while directly inheriting the preceding context's retained state assumes that its knowledge remains relevant to the new environment.
When consecutive driving conditions differ substantially, this assumption may lead to unstable adaptation.
KROMA therefore evaluates previously accepted states on current observations and selects a suitable historical initialization based on detection loss, rather than assuming that consecutive contexts share transferable knowledge.

Specifically, we construct a scoring subset $\mathcal{P}_t \subseteq \mathcal{B}_t$ from detection-eligible observations assigned to online learning.
Before adapting to the new context, KROMA evaluates previously accepted detector states on the same observations and selects the one with the lowest average detection loss:
\begin{equation}
    c_t^* = \arg\min_{c \in \mathcal{C}_{t-1}}
    \frac{1}{|\mathcal{P}_t|}
    \sum_{(x_i,y_i)\in\mathcal{P}_t}
    \mathcal{L}_{\mathrm{det}}(f_c^{\mathrm{A}}(x_i),y_i).
    \label{eq:historical_transfer}
\end{equation}
where $\mathcal{C}_{t-1}$ denotes the historical contexts stored in $\mathcal{M}_{t-1}$, $f_c^{\mathrm{A}}$ is the accepted detector of context $c$, and $\mathcal{L}_{\mathrm{det}}$ denotes the primary detection loss.
The selected state provides the starting point for the new context, providing the foundation for subsequent knowledge-preserving adaptation.

\subsection{Knowledge-Preserving Online Adaptation}
\label{sec:knowledge_preserving}

\noindent\textbf{Context-Specific Residual Adaptation.}
Continuously updating shared parameters can interfere with knowledge acquired under previously encountered driving conditions.
To preserve a common foundation while accommodating diverse driving conditions, KROMA maintains frozen shared parameters and learns context-specific modifications independently.
Specifically, each context $c$ uses fixed binary sign codes $\mathbf{s}_c$ to modulate shared weights, while trainable residuals $\Delta_c$ capture context-specific adaptations.
For each sign-modulated layer $\ell$, the effective weights are defined as
\begin{equation}
    \widetilde{\Theta}_{c,\ell}
    =
    \Theta_\ell \odot S_{c,\ell}
    +
    \mathcal{T}_{S_{c,\ell}}(\Delta_{c,\ell}),
    \label{eq:psp_effective}
\end{equation}
where $\Theta_\ell$ denotes the frozen shared weights, $S_{c,\ell}\in\{-1,+1\}$ is the layer-specific sign code within $\mathbf{s}_c=\{S_{c,\ell}\}_{\ell}$, and $\mathcal{T}_{S_{c,\ell}}$ maps residuals into the effective weight space.
This mapping applies sign modulation to linear-projection residuals and leaves convolutional residuals unchanged, making it invertible under fixed binary codes.
During online adaptation, only the active context's residuals in the fusion module and detection head are updated, while the shared parameters and all other context-specific residuals remain frozen.

When adapting to a new context, KROMA inherits the effective parameters of the selected historical state by initializing the corresponding residuals under the new sign code.
Let $n=c_{\mathrm{new}}$ denote the new context and $p=c_t^*$ its selected historical parent.
The initial residuals for each sign-modulated layer $\ell$ are computed as
\begin{equation}
    \Delta_{n,\ell}^{(0)}
    =
    \mathcal{T}_{S_{n,\ell}}^{-1}
    \left(
        \widetilde{\Theta}_{p,\ell}^{\mathrm{A}}
        -
        \Theta_\ell\odot S_{n,\ell}
    \right),
    \label{eq:psp_inheritance}
\end{equation}
where $\widetilde{\Theta}_{p,\ell}^{\mathrm{A}}$ denotes the effective weights of the selected historical accepted state.
The associated normalization codes, bias residuals, and fusion-query residuals are also inherited, preserving the selected detector's function at initialization:
$f_n^{(0)}(x)=f_p^{\mathrm{A}}(x)$.
The new context can then adapt independently, with $\Delta_n^{(0)}$ serving as a fixed reference for subsequent knowledge-constrained learning.

\noindent\textbf{Knowledge-Constrained Online Learning.}
Although context-specific residuals prevent direct interference across environments, successive online updates may still degrade knowledge inherited from historical states or acquired earlier within the current context.
To preserve inherited knowledge, KROMA constrains the learner residuals $\Delta_c^{\mathrm{L}}$ toward their initial values $\Delta_c^{(0)}$, which serve as a fixed reference during adaptation.
Meanwhile, a bounded context-specific rehearsal memory $\mathcal{H}_c$ retains previously trained observations to preserve knowledge acquired earlier within the current context.

For each observation $(x_t,y_t)$ selected for online training, KROMA applies the complete detection loss to the full-fusion branch and classification supervision to the single- and dual-modality auxiliary branches.
The current supervision is defined as
\begin{equation}
\begin{aligned}
    \mathcal{L}_{\mathrm{cur},t}
    &= \mathcal{L}_{\mathrm{det}}(f_c^{\mathrm{L}}(x_t),y_t) \\
    &\quad + \frac{\alpha}{|\mathcal{U}|}
    \sum_{u\in\mathcal{U}}
    \mathcal{L}_{\mathrm{cls}}(f_{c,u}^{\mathrm{L}}(x_t),y_t),
\end{aligned}
\label{eq:current_supervision}
\end{equation}
where $f_c^{\mathrm{L}}$ denotes the detector using the current learner residuals $\Delta_c^{\mathrm{L}}$, $\mathcal{U}$ denotes the set of auxiliary sensor branches, $f_{c,u}^{\mathrm{L}}$ denotes the corresponding branch output, and $\alpha$ controls the contribution of auxiliary supervision.

To retain previously acquired experience, a rehearsal observation $(x_h,y_h)$ is sampled from $\mathcal{H}_c$ when available, with $\mathcal{L}_{\mathrm{rep},t}=\mathcal{L}_{\mathrm{det}}(f_c^{\mathrm{L}}(x_h),y_h)$.
Combining current supervision, context-specific rehearsal, and regularization toward the inherited state, we formulate the online objective as
\begin{equation}
    \mathcal{L}_t =
    \frac{\mathcal{L}_{\mathrm{cur},t}
    + \beta_t \mathcal{L}_{\mathrm{rep},t}}
    {1+\beta_t}
    + \frac{\lambda}{2}
    \left\|
        \Delta_c^{\mathrm{L}}-\Delta_c^{(0)}
    \right\|_2^2,
    \label{eq:knowledge_constrained}
\end{equation}
where $\beta_t=\beta$ when a rehearsal sample is available and zero otherwise, and $\lambda$ controls the inherited-reference constraint.
Since the shared parameters and sign codes remain frozen, regularizing the residuals also limits deviations of the effective weights from their inherited values.
Together with context-specific rehearsal, this allows the learner to adapt to new observations while helping preserve inherited knowledge and previously acquired capabilities.

\noindent\textbf{Selective State Retention.}
Although regularization and context-specific rehearsal help preserve acquired knowledge, they cannot guarantee that the latest model state provides the best detection performance.
KROMA therefore separates ongoing learning from knowledge retention by maintaining two complementary states for each context: a \emph{learner state} that continuously adapts to incoming observations and an \emph{accepted state} that retains a performance-selected parameter state for future reuse.
To guide this selection, we maintain a bounded context-specific pool $\mathcal{V}_c$ of labeled observations, using class-aware sampling to improve coverage across object categories.
These observations are temporarily reserved for model selection, allowing KROMA to compare the detection performance of different candidate states on the same samples.

During online adaptation, KROMA periodically performs this comparison among the learner and accepted states, together with the initial inherited state when distinct.
The state with the highest class-averaged detection AP across multiple IoU thresholds is selected:
\begin{equation}
    q_c^* = \arg\max_{q \in \mathcal{Q}_c}
    \mathrm{AP}_{\mathrm{sel}}
    \left(f_c^{(q)};\mathcal{V}_c\right),
    \label{eq:state_selection}
\end{equation}
where $\mathcal{Q}_c \subseteq \{\mathrm{L},\mathrm{A},0\}$ denotes the candidate set of learner, accepted, and inherited states, and $f_c^{(q)}$ represents the detector under candidate state $q$.
$\mathrm{AP}_{\mathrm{sel}}$ averages detection AP over represented classes and selected IoU thresholds, using the primary detection loss to break ties.
The accepted state is updated only when a better candidate is identified, while the learner continues adapting independently.
These accepted states are retained in the semantic context memory $\mathcal{M}_t$, allowing future adaptation to build on knowledge acquired under previously encountered driving conditions.


\begin{table*}[t]
\centering
\caption{
Overall performance on K-Radar (mean ± std. over five random context orders). Target and All denote macro-averages over nine target and all ten contexts, respectively. The \textbf{best} and \underline{second-best} mean AP values are shown \textbf{bold} and \underline{underlined}.
}
\label{tab:overall_performance}

\footnotesize
\setlength{\tabcolsep}{3.7pt}
\renewcommand{\arraystretch}{1.00}

\newcommand{\APstrut}{\rule[-0.4ex]{0pt}{2ex}}

\setbox0=\hbox{\textbf{00}}
\edef\APdigitswidth{\the\wd0}
\setbox0=\hbox{\textbf{.}}
\edef\APpointwidth{\the\wd0}
\newcommand{\APnumber}[3][]{%
  #1{\makebox[\APdigitswidth][r]{#2}%
     \makebox[\APpointwidth][c]{.}%
     \makebox[\APdigitswidth][l]{#3}}%
}

\newcommand{\APpm}{%
  \hspace{0.22em}%
  \shortstack{\APstrut$\pm$\\[-0.6pt]\APstrut$\pm$}%
  \hspace{0.22em}%
}
\begin{tabular}{@{}
  >{\hspace{\tabcolsep}}l<{\hspace{\tabcolsep}}@{}
  >{\hspace{\tabcolsep}}l<{\hspace{\tabcolsep}}@{}
  >{\hspace{\tabcolsep}}c<{\hspace{\tabcolsep}}@{}
  *{6}{>{\hspace{\tabcolsep}}r@{}c@{}l<{\hspace{\tabcolsep}}@{}}
}
\toprule

\multirow{2}{*}{Method} &
\multirow{2}{*}{Venue} &
\multirow{2}{*}{IoU} &
\multicolumn{6}{c}{Source Context} &
\multicolumn{6}{c}{Target Contexts} &
\multicolumn{6}{c}{All Contexts} \\

\cmidrule(lr){4-9}
\cmidrule(lr){10-15}
\cmidrule(lr){16-21}

& & &
\multicolumn{3}{c}{$\mathrm{AP}_{\mathrm{BEV}}\uparrow$ (\%)}&
\multicolumn{3}{c}{$\mathrm{AP}_{3D}\uparrow$ (\%)}&
\multicolumn{3}{c}{$\mathrm{AP}_{\mathrm{BEV}}\uparrow$ (\%)}&
\multicolumn{3}{c}{$\mathrm{AP}_{3D}\uparrow$ (\%)}&
\multicolumn{3}{c}{$\mathrm{AP}_{\mathrm{BEV}}\uparrow$ (\%)}&
\multicolumn{3}{c}{$\mathrm{AP}_{3D}\uparrow$ (\%)}\\

\midrule


\rowcolor{black!3}[0pt][0pt]
Source-only & -- & \shortstack{\APstrut0.3\\[-0.6pt]\APstrut0.5} &
\shortstack{\APstrut\APnumber{68}{87}\\[-0.6pt]\APstrut\APnumber{66}{43}} & \APpm & \shortstack{\APstrut\APnumber{0}{00}\\[-0.6pt]\APstrut\APnumber{0}{00}} & \shortstack{\APstrut\APnumber[\underline]{68}{60}\\[-0.6pt]\APstrut\APnumber[\underline]{50}{30}} & \APpm & \shortstack{\APstrut\APnumber{0}{00}\\[-0.6pt]\APstrut\APnumber{0}{00}} &
\shortstack{\APstrut\APnumber{73}{47}\\[-0.6pt]\APstrut\APnumber{42}{87}} & \APpm & \shortstack{\APstrut\APnumber{0}{00}\\[-0.6pt]\APstrut\APnumber{0}{00}} & \shortstack{\APstrut\APnumber{54}{47}\\[-0.6pt]\APstrut\APnumber{15}{57}} & \APpm & \shortstack{\APstrut\APnumber{0}{00}\\[-0.6pt]\APstrut\APnumber{0}{00}} &
\shortstack{\APstrut\APnumber{73}{01}\\[-0.6pt]\APstrut\APnumber{45}{23}} & \APpm & \shortstack{\APstrut\APnumber{0}{00}\\[-0.6pt]\APstrut\APnumber{0}{00}} & \shortstack{\APstrut\APnumber{55}{89}\\[-0.6pt]\APstrut\APnumber{19}{04}} & \APpm & \shortstack{\APstrut\APnumber{0}{00}\\[-0.6pt]\APstrut\APnumber{0}{00}} \\

\midrule


CTAOD~\cite{yoo2024and} & CVPR'24 & \shortstack{\APstrut0.3\\[-0.6pt]\APstrut0.5} &
\shortstack{\APstrut\APnumber[\underline]{71}{21}\\[-0.6pt]\APstrut\APnumber{61}{72}} & \APpm & \shortstack{\APstrut\APnumber{0}{81}\\[-0.6pt]\APstrut\APnumber{11}{00}} & \shortstack{\APstrut\APnumber{54}{36}\\[-0.6pt]\APstrut\APnumber{11}{22}} & \APpm & \shortstack{\APstrut\APnumber{26}{73}\\[-0.6pt]\APstrut\APnumber{11}{85}} &
\shortstack{\APstrut\APnumber[\underline]{74}{96}\\[-0.6pt]\APstrut\APnumber{60}{95}} & \APpm & \shortstack{\APstrut\APnumber{1}{73}\\[-0.6pt]\APstrut\APnumber{3}{46}} & \shortstack{\APstrut\APnumber{66}{59}\\[-0.6pt]\APstrut\APnumber{25}{14}} & \APpm & \shortstack{\APstrut\APnumber{8}{40}\\[-0.6pt]\APstrut\APnumber{14}{66}} &
\shortstack{\APstrut\APnumber[\underline]{74}{58}\\[-0.6pt]\APstrut\APnumber{61}{03}} & \APpm & \shortstack{\APstrut\APnumber{1}{49}\\[-0.6pt]\APstrut\APnumber{4}{02}} & \shortstack{\APstrut\APnumber{65}{37}\\[-0.6pt]\APstrut\APnumber{23}{75}} & \APpm & \shortstack{\APstrut\APnumber{10}{19}\\[-0.6pt]\APstrut\APnumber{13}{45}} \\

\addlinespace[0pt]

\rowcolor{black!3}[0pt][0pt]
BECoTTA~\cite{lee2024becotta} & ICML'24 & \shortstack{\APstrut0.3\\[-0.6pt]\APstrut0.5} &
\shortstack{\APstrut\APnumber{70}{45}\\[-0.6pt]\APstrut\APnumber{67}{05}} & \APpm & \shortstack{\APstrut\APnumber{0}{89}\\[-0.6pt]\APstrut\APnumber{1}{12}} & \shortstack{\APstrut\APnumber[\textbf]{68}{74}\\[-0.6pt]\APstrut\APnumber[\textbf]{50}{85}} & \APpm & \shortstack{\APstrut\APnumber{0}{61}\\[-0.6pt]\APstrut\APnumber{2}{65}} &
\shortstack{\APstrut\APnumber{73}{92}\\[-0.6pt]\APstrut\APnumber{53}{47}} & \APpm & \shortstack{\APstrut\APnumber{0}{71}\\[-0.6pt]\APstrut\APnumber{3}{70}} & \shortstack{\APstrut\APnumber{65}{48}\\[-0.6pt]\APstrut\APnumber{22}{17}} & \APpm & \shortstack{\APstrut\APnumber{2}{60}\\[-0.6pt]\APstrut\APnumber{4}{78}} &
\shortstack{\APstrut\APnumber{73}{58}\\[-0.6pt]\APstrut\APnumber{54}{83}} & \APpm & \shortstack{\APstrut\APnumber{0}{69}\\[-0.6pt]\APstrut\APnumber{3}{43}} & \shortstack{\APstrut\APnumber{65}{80}\\[-0.6pt]\APstrut\APnumber{25}{04}} & \APpm & \shortstack{\APstrut\APnumber{2}{36}\\[-0.6pt]\APstrut\APnumber{4}{16}} \\

\addlinespace[0pt]

DPO~\cite{chen2024dpo} & MM'24 & \shortstack{\APstrut0.3\\[-0.6pt]\APstrut0.5} &
\shortstack{\APstrut\APnumber{69}{38}\\[-0.6pt]\APstrut\APnumber{61}{32}} & \APpm & \shortstack{\APstrut\APnumber{4}{01}\\[-0.6pt]\APstrut\APnumber{4}{45}} & \shortstack{\APstrut\APnumber{54}{72}\\[-0.6pt]\APstrut\APnumber{14}{47}} & \APpm & \shortstack{\APstrut\APnumber{23}{69}\\[-0.6pt]\APstrut\APnumber{12}{40}} &
\shortstack{\APstrut\APnumber{61}{94}\\[-0.6pt]\APstrut\APnumber{52}{54}} & \APpm & \shortstack{\APstrut\APnumber{11}{40}\\[-0.6pt]\APstrut\APnumber{7}{28}} & \shortstack{\APstrut\APnumber{54}{52}\\[-0.6pt]\APstrut\APnumber{26}{54}} & \APpm & \shortstack{\APstrut\APnumber{9}{74}\\[-0.6pt]\APstrut\APnumber{14}{14}} &
\shortstack{\APstrut\APnumber{62}{68}\\[-0.6pt]\APstrut\APnumber{53}{42}} & \APpm & \shortstack{\APstrut\APnumber{9}{99}\\[-0.6pt]\APstrut\APnumber{6}{59}} & \shortstack{\APstrut\APnumber{54}{54}\\[-0.6pt]\APstrut\APnumber{25}{33}} & \APpm & \shortstack{\APstrut\APnumber{10}{06}\\[-0.6pt]\APstrut\APnumber{13}{11}} \\

\addlinespace[0pt]


\rowcolor{black!3}[0pt][0pt]
MOS~\cite{chen2025mos} & ICLR'25 & \shortstack{\APstrut0.3\\[-0.6pt]\APstrut0.5} &
\shortstack{\APstrut\APnumber{70}{24}\\[-0.6pt]\APstrut\APnumber{57}{80}} & \APpm & \shortstack{\APstrut\APnumber{1}{10}\\[-0.6pt]\APstrut\APnumber{15}{72}} & \shortstack{\APstrut\APnumber{54}{89}\\[-0.6pt]\APstrut\APnumber{15}{95}} & \APpm & \shortstack{\APstrut\APnumber{19}{48}\\[-0.6pt]\APstrut\APnumber{19}{93}} &
\shortstack{\APstrut\APnumber{73}{55}\\[-0.6pt]\APstrut\APnumber{60}{00}} & \APpm & \shortstack{\APstrut\APnumber{6}{20}\\[-0.6pt]\APstrut\APnumber{4}{94}} & \shortstack{\APstrut\APnumber[\underline]{67}{11}\\[-0.6pt]\APstrut\APnumber{28}{31}} & \APpm & \shortstack{\APstrut\APnumber{4}{51}\\[-0.6pt]\APstrut\APnumber{15}{22}} &
\shortstack{\APstrut\APnumber{73}{22}\\[-0.6pt]\APstrut\APnumber{59}{78}} & \APpm & \shortstack{\APstrut\APnumber{5}{50}\\[-0.6pt]\APstrut\APnumber{5}{28}} & \shortstack{\APstrut\APnumber[\underline]{65}{88}\\[-0.6pt]\APstrut\APnumber{27}{08}} & \APpm & \shortstack{\APstrut\APnumber{5}{48}\\[-0.6pt]\APstrut\APnumber{13}{79}} \\

\addlinespace[0pt]

Hybrid-TTA~\cite{park2025hybrid} & ICCV'25 & \shortstack{\APstrut0.3\\[-0.6pt]\APstrut0.5} &
\shortstack{\APstrut\APnumber{69}{60}\\[-0.6pt]\APstrut\APnumber[\underline]{67}{18}} & \APpm & \shortstack{\APstrut\APnumber{1}{08}\\[-0.6pt]\APstrut\APnumber{1}{07}} & \shortstack{\APstrut\APnumber{65}{86}\\[-0.6pt]\APstrut\APnumber{28}{29}} & \APpm & \shortstack{\APstrut\APnumber{2}{02}\\[-0.6pt]\APstrut\APnumber{6}{44}} &
\shortstack{\APstrut\APnumber{67}{07}\\[-0.6pt]\APstrut\APnumber[\underline]{62}{43}} & \APpm & \shortstack{\APstrut\APnumber{9}{37}\\[-0.6pt]\APstrut\APnumber{8}{33}} & \shortstack{\APstrut\APnumber{65}{76}\\[-0.6pt]\APstrut\APnumber[\underline]{39}{85}} & \APpm & \shortstack{\APstrut\APnumber{9}{05}\\[-0.6pt]\APstrut\APnumber{4}{04}} &
\shortstack{\APstrut\APnumber{67}{32}\\[-0.6pt]\APstrut\APnumber[\underline]{62}{91}} & \APpm & \shortstack{\APstrut\APnumber{8}{38}\\[-0.6pt]\APstrut\APnumber{7}{44}} & \shortstack{\APstrut\APnumber{65}{77}\\[-0.6pt]\APstrut\APnumber[\underline]{38}{70}} & \APpm & \shortstack{\APstrut\APnumber{8}{16}\\[-0.6pt]\APstrut\APnumber{3}{66}} \\

\addlinespace[0pt]

\rowcolor{black!3}[0pt][0pt]
CodeMerge~\cite{yang2025codemerge} & NeurIPS'25 & \shortstack{\APstrut0.3\\[-0.6pt]\APstrut0.5} &
\shortstack{\APstrut\APnumber{69}{33}\\[-0.6pt]\APstrut\APnumber{55}{79}} & \APpm & \shortstack{\APstrut\APnumber{1}{64}\\[-0.6pt]\APstrut\APnumber{17}{38}} & \shortstack{\APstrut\APnumber{52}{31}\\[-0.6pt]\APstrut\APnumber{11}{57}} & \APpm & \shortstack{\APstrut\APnumber{18}{70}\\[-0.6pt]\APstrut\APnumber{13}{50}} &
\shortstack{\APstrut\APnumber{69}{15}\\[-0.6pt]\APstrut\APnumber{57}{46}} & \APpm & \shortstack{\APstrut\APnumber{8}{81}\\[-0.6pt]\APstrut\APnumber{4}{67}} & \shortstack{\APstrut\APnumber{60}{84}\\[-0.6pt]\APstrut\APnumber{24}{24}} & \APpm & \shortstack{\APstrut\APnumber{3}{78}\\[-0.6pt]\APstrut\APnumber{13}{39}} &
\shortstack{\APstrut\APnumber{69}{17}\\[-0.6pt]\APstrut\APnumber{57}{29}} & \APpm & \shortstack{\APstrut\APnumber{7}{85}\\[-0.6pt]\APstrut\APnumber{5}{57}} & \shortstack{\APstrut\APnumber{59}{98}\\[-0.6pt]\APstrut\APnumber{22}{98}} & \APpm & \shortstack{\APstrut\APnumber{4}{95}\\[-0.6pt]\APstrut\APnumber{12}{41}} \\

\addlinespace[0pt]


AMROD~\cite{cao2026continual} & ESWA'26 & \shortstack{\APstrut0.3\\[-0.6pt]\APstrut0.5} &
\shortstack{\APstrut\APnumber[\textbf]{71}{60}\\[-0.6pt]\APstrut\APnumber[\textbf]{68}{23}} & \APpm & \shortstack{\APstrut\APnumber{1}{09}\\[-0.6pt]\APstrut\APnumber{0}{33}} & \shortstack{\APstrut\APnumber{67}{74}\\[-0.6pt]\APstrut\APnumber{33}{93}} & \APpm & \shortstack{\APstrut\APnumber{2}{17}\\[-0.6pt]\APstrut\APnumber{9}{03}} &
\shortstack{\APstrut\APnumber{58}{81}\\[-0.6pt]\APstrut\APnumber{47}{73}} & \APpm & \shortstack{\APstrut\APnumber{12}{95}\\[-0.6pt]\APstrut\APnumber{5}{62}} & \shortstack{\APstrut\APnumber{54}{73}\\[-0.6pt]\APstrut\APnumber{25}{20}} & \APpm & \shortstack{\APstrut\APnumber{11}{06}\\[-0.6pt]\APstrut\APnumber{4}{33}} &
\shortstack{\APstrut\APnumber{60}{09}\\[-0.6pt]\APstrut\APnumber{49}{78}} & \APpm & \shortstack{\APstrut\APnumber{11}{69}\\[-0.6pt]\APstrut\APnumber{5}{02}} & \shortstack{\APstrut\APnumber{56}{03}\\[-0.6pt]\APstrut\APnumber{26}{08}} & \APpm & \shortstack{\APstrut\APnumber{10}{01}\\[-0.6pt]\APstrut\APnumber{4}{38}} \\

\midrule


\rowcolor{black!3}[0pt][0pt]
\textbf{KROMA (Ours)} & -- & \shortstack{\APstrut0.3\\[-0.6pt]\APstrut0.5} &
\shortstack{\APstrut\APnumber{68}{87}\\[-0.6pt]\APstrut\APnumber{66}{43}} & \APpm & \shortstack{\APstrut\APnumber{0}{00}\\[-0.6pt]\APstrut\APnumber{0}{00}} & \shortstack{\APstrut\APnumber[\underline]{68}{60}\\[-0.6pt]\APstrut\APnumber[\underline]{50}{30}} & \APpm & \shortstack{\APstrut\APnumber{0}{00}\\[-0.6pt]\APstrut\APnumber{0}{00}} &
\shortstack{\APstrut\APnumber[\textbf]{76}{50}\\[-0.6pt]\APstrut\APnumber[\textbf]{65}{33}} & \APpm & \shortstack{\APstrut\APnumber{0}{55}\\[-0.6pt]\APstrut\APnumber{1}{89}} & \shortstack{\APstrut\APnumber[\textbf]{74}{10}\\[-0.6pt]\APstrut\APnumber[\textbf]{48}{44}} & \APpm & \shortstack{\APstrut\APnumber{1}{34}\\[-0.6pt]\APstrut\APnumber{1}{33}} &
\shortstack{\APstrut\APnumber[\textbf]{75}{74}\\[-0.6pt]\APstrut\APnumber[\textbf]{65}{44}} & \APpm & \shortstack{\APstrut\APnumber{0}{49}\\[-0.6pt]\APstrut\APnumber{1}{70}} & \shortstack{\APstrut\APnumber[\textbf]{73}{55}\\[-0.6pt]\APstrut\APnumber[\textbf]{48}{62}} & \APpm & \shortstack{\APstrut\APnumber{1}{21}\\[-0.6pt]\APstrut\APnumber{1}{20}} \\

\bottomrule

\end{tabular}
\end{table*}

\section{Experiments}
\label{sec:experiments}

\subsection{Experimental Setup}
\label{sec:experimental_setup}

\noindent\textbf{Evaluation Protocols.}
We evaluate KROMA on K-Radar~\cite{paek2022k}, as it provides a large-scale real-world 3D object detection dataset with synchronized camera, LiDAR, and 4D radar measurements and 3D annotations across diverse combinations of weather, road types, and lighting conditions.
Each sequence is chronologically split into 60\% for learning and 40\% for evaluation.
The source detector is pretrained offline on Sequence 1, while the remaining nine sequences are processed as an online stream, with supervised updates performed as frames arrive in a single pass rather than through multi-epoch offline training.
Our main comparative experiments evaluate all methods across five randomly sampled context orders, with identical data splits and a shared source-pretrained checkpoint. Results are reported as mean $\pm$ standard deviation where applicable. Additional experiments follow the specific evaluation protocols described in their respective sections.

\noindent\textbf{Baselines.}
We compare KROMA against seven representative online adaptation methods.
MOS~\cite{chen2025mos} and CodeMerge~\cite{yang2025codemerge} exploit historical checkpoints for knowledge reuse, while BECoTTA~\cite{lee2024becotta} employs domain-adaptive expert routing.
Hybrid-TTA~\cite{park2025hybrid}, CTAOD~\cite{yoo2024and}, AMROD~\cite{cao2026continual}, and DPO~\cite{chen2024dpo} address adaptation robustness through dynamic updating, parameter restoration, or perturbation-based optimization.

\noindent\textbf{Metrics.}
We evaluate detection performance using bird's-eye-view (BEV) and 3D average precision ($\mathrm{AP}_{\mathrm{BEV}}^{\theta}$ and $\mathrm{AP}_{\mathrm{3D}}^{\theta}$, respectively), where $\theta \in \{0.3, 0.5\}$ denotes the IoU threshold.
Results are macro-averaged across contexts within each evaluation group, while mean $\mathrm{AP}_{\mathrm{BEV}}$ and $\mathrm{AP}_{\mathrm{3D}}$ average the scores over both IoU thresholds.
We measure computational efficiency using FPS, accounting for both online adaptation and subsequent target inference.

\noindent\textbf{Implementation Details.}
We adopt Availability-aware Sensor Fusion (ASF)~\cite{paek2026availability} as the camera--LiDAR--radar detection backbone.
All methods are initialized from the same checkpoint and follow a unified supervised online adaptation protocol for fair comparison.
The attribute predictor is pretrained on a small set of 300 frames from the training splits, using only weather, road-type, and lighting labels without explicit scene identities and 3D detection annotations.
During online adaptation, KROMA uses a five-frame context decision window, a confidence threshold of 0.9, and a candidate pool with a capacity of 32.
These hyperparameters are selected using validation data and kept fixed across all experiments, unless otherwise specified.
Model updates use AdamW with a learning rate of $10^{-4}$.
All experiments are conducted on a single NVIDIA A100 GPU.

\subsection{Overall Performance}
\label{sec:overall_performance}

As shown in Table~\ref{tab:overall_performance}, KROMA achieves the highest $\mathrm{AP}_{\mathrm{BEV}}$ and $\mathrm{AP}_{\mathrm{3D}}$ at both IoU thresholds across all contexts.
It achieves 73.55\% $\mathrm{AP}_{\mathrm{3D}}^{0.3}$ and 48.62\% $\mathrm{AP}_{\mathrm{3D}}^{0.5}$, outperforming the strongest baseline at each threshold by 7.67 and 9.92 percentage points (pp), respectively.
Unlike BEV AP, which measures overlap on the ground plane, 3D AP also accounts for vertical alignment and object height, which are important for accurate localization in practical autonomous driving.
The larger gain at IoU 0.5 highlights KROMA's advantage under stricter 3D localization requirements.

CTAOD~\cite{yoo2024and}, DPO~\cite{chen2024dpo}, Hybrid-TTA~\cite{park2025hybrid}, and AMROD~\cite{cao2026continual} improve adaptation through update scheduling, perturbation-based optimization, adaptive tuning, and parameter restoration, but do not explicitly select context-relevant historical states for initialization.
KROMA instead retrieves suitable historical knowledge to guide adaptation under new driving conditions.
Across the nine target contexts, KROMA achieves 48.44\% $\mathrm{AP}_{\mathrm{3D}}^{0.5}$, outperforming these methods by 8.59--23.30 pp.
MOS~\cite{chen2025mos} and CodeMerge~\cite{yang2025codemerge} exploit historical checkpoints through model selection and merging, while BECoTTA~\cite{lee2024becotta} employs domain-adaptive expert routing.
Compared with these methods, KROMA further combines context-aware historical retrieval with context-specific residual updates, knowledge constraints, and selective state retention.
Compared with MOS, the strongest among these baselines, KROMA improves target-context $\mathrm{AP}_{\mathrm{3D}}^{0.5}$ by 20.13 pp.
These results support the effectiveness of combining context-aware knowledge reuse with knowledge-preserving online adaptation across changing driving conditions.

\begin{table*}[t]
\centering
\caption{\textbf{Performance and knowledge retention under recurring driving contexts.} Mean $\mathrm{AP}_{\mathrm{3D}}$ (\%) is averaged over IoU thresholds of 0.3 and 0.5 and across nine target contexts.
All methods follow the same randomly sampled context order across three rounds.
Start and End denote detection performance before and after each visit, respectively. $F_2$ and $F_3$ measure inter-visit forgetting (pp, lower is better). The best and second-best results are highlighted in \textbf{bold} and \underline{underlined}, respectively.}

\label{tab:sup_recurrence}

\footnotesize
\setlength{\tabcolsep}{5pt}
\renewcommand{\arraystretch}{1.0}

\newcommand{\RecVal}[2][]{%
    \makebox[4.0em][c]{#1{#2}}%
}

\begin{tabular}{@{}ll*{7}{c}@{}}
\toprule

\multirow{2}{*}{Method} &
\multirow{2}{*}{Venue} &
\multicolumn{1}{c}{Visit 1} &
\multicolumn{2}{c}{Visit 2} &
\multicolumn{2}{c}{Visit 3} &
\multicolumn{2}{c}{Forgetting} \\

\cmidrule(lr){3-3}
\cmidrule(lr){4-5}
\cmidrule(lr){6-7}
\cmidrule(lr){8-9}

& &
End $\uparrow$ &
Start $\uparrow$ & End $\uparrow$ &
Start $\uparrow$ & End $\uparrow$ &
$F_2 \downarrow$ & $F_3 \downarrow$ \\

\midrule


CTAOD~\cite{yoo2024and} & CVPR'24 &
\RecVal{43.33} & \RecVal{40.66} & \RecVal[\underline]{52.74} &
\RecVal{38.85} & \RecVal{55.94} &
\RecVal{11.61} & \RecVal{17.52} \\

\rowcolor{black!3}
BECoTTA~\cite{lee2024becotta} & ICML'24 &
\RecVal{37.75} & \RecVal{41.47} & \RecVal{42.01} &
\RecVal{43.42} & \RecVal{44.47} &
\RecVal[\underline]{0.23} & \RecVal[\underline]{0.34} \\

DPO~\cite{chen2024dpo} & MM'24 &
\RecVal{43.18} & \RecVal{34.02} & \RecVal{49.92} &
\RecVal{39.51} & \RecVal{54.38} &
\RecVal{14.48} & \RecVal{14.57} \\


\rowcolor{black!3}
MOS~\cite{chen2025mos} & ICLR'25 &
\RecVal[\underline]{48.02} & \RecVal{45.78} & \RecVal{52.45} &
\RecVal{47.12} & \RecVal[\underline]{56.25} &
\RecVal{7.50} & \RecVal{11.42} \\

Hybrid-TTA~\cite{park2025hybrid} & ICCV'25 &
\RecVal{42.61} & \RecVal[\underline]{47.95} & \RecVal{49.98} &
\RecVal[\underline]{51.61} & \RecVal{52.85} &
\RecVal{0.67} & \RecVal{2.02} \\

\rowcolor{black!3}
CodeMerge~\cite{yang2025codemerge} & NeurIPS'25 &
\RecVal{47.12} & \RecVal{33.41} & \RecVal{39.46} &
\RecVal{36.67} & \RecVal{54.66} &
\RecVal{14.56} & \RecVal{10.08} \\


AMROD~\cite{cao2026continual} & ESWA'26 &
\RecVal{37.60} & \RecVal{40.60} & \RecVal{42.01} &
\RecVal{37.56} & \RecVal{37.47} &
\RecVal{1.81} & \RecVal{9.00} \\

\midrule


\rowcolor{black!3}
\textbf{KROMA (Ours)} & -- &
\RecVal[\textbf]{51.28} &
\RecVal[\textbf]{51.28} & \RecVal[\textbf]{53.31} &
\RecVal[\textbf]{53.31} & \RecVal[\textbf]{58.25} &
\RecVal[\textbf]{0.00} & \RecVal[\textbf]{0.00} \\

\bottomrule
\end{tabular}
\end{table*}

\begin{figure*}[t]
    \centering

    \begin{minipage}[t]{0.65\textwidth}
        \vspace{0pt}
        \centering
        \includegraphics[
            width=\linewidth
        ]{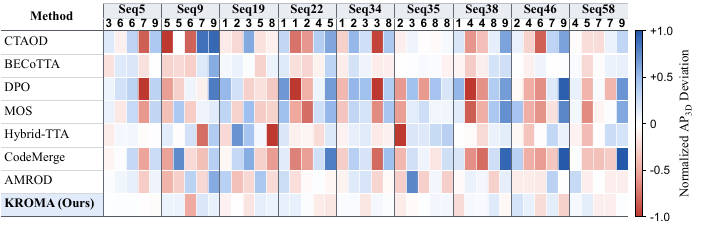}\label{fig:arrival_heatmap}

        \vspace{2pt}
        {\footnotesize\textbf{(a)} Context Arrival Position Across Five Orders\par}
    \end{minipage}%
    \hfill
    \begin{minipage}[t]{0.33\textwidth}
        \vspace{0pt}
        \centering
        \includegraphics[
            width=\linewidth
        ]{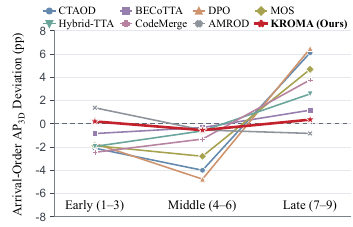}\label{fig:arrival_stage}

        \vspace{2pt}
        {\footnotesize\textbf{(b)} Context Arrival Stage\par}
    \end{minipage}

    \vspace{3pt}
    \caption{
        \textbf{Forgetting-Related Order Sensitivity.}
Deviations are computed by subtracting each method--context pair's
mean $\mathrm{AP}_{\mathrm{3D}}$ across five arrival orders.
(a) Per-context deviations normalized using a scale shared
across methods within each context and sorted by arrival position.
(b) Unnormalized deviations (pp) averaged across
Early (1--3), Middle (4--6), and Late (7--9) arrival stages.  }
    \label{fig:arrival_order}
\end{figure*}

\subsection{Knowledge Retention and Stability Analysis}
\label{sec:knowledge_retention}

Knowledge retention is essential for continual adaptation, as adapting a detector to new driving conditions can degrade its performance in previously encountered environments.
To assess this, we first average $\mathrm{AP}_{\mathrm{3D}}^{0.3}$ and $\mathrm{AP}_{\mathrm{3D}}^{0.5}$ for each method, context, and order, then subtract the mean for the same method and context across all five orders.
Fig.~\ref{fig:arrival_order}(a) visualizes these deviations by context arrival position, normalized using a scale shared across methods within each context.
Blue and red indicate above- and below-average performance, respectively, while colors closer to white indicate smaller deviations.
Fig.~\ref{fig:arrival_order}(b) further groups the unnormalized deviations into Early (1--3), Middle (4--6), and Late (7--9) stages to examine the effects of arrival position.

The baselines exhibit substantial fluctuations in AP$_{\mathrm{3D}}$ across context arrival orders, with an overall mean absolute deviation of 9.67~pp, particularly for DPO~\cite{chen2024dpo}, CodeMerge~\cite{yang2025codemerge}, and CTAOD~\cite{yoo2024and}.
In contrast, KROMA achieves the lowest mean absolute deviation of 2.95 pp, indicating more consistent detection performance across different context orders.
Most baselines show negative deviations for earlier contexts and positive deviations for later ones, suggesting that final performance depends on when a context is encountered, reflecting interference from subsequent adaptation.
For example, the ranges between the maximum and minimum stage-wise deviations reach 11.23 pp for DPO and 10.11 pp for CTAOD.
By comparison, KROMA's Early, Middle, and Late deviations are +0.20, -0.55, and +0.35 pp, respectively, with a range of only 0.90 pp.
These results provide further evidence that KROMA effectively preserves acquired knowledge and maintains stability while continuing to adapt to new driving conditions.

\subsection{Context Recurrence \& Knowledge Reactivation}
\label{sec:sup_recurrence}

In real-world driving, previously encountered conditions may recur after the detector has adapted to other environments. To evaluate whether KROMA can reliably reactivate historical knowledge and continue adapting when these conditions reappear, we conduct a context recurrence and knowledge reactivation experiment. Specifically, we divide the original 60\% split used for online adaptation in each target sequence into three disjoint chronological segments, while keeping the 40\% test split unchanged. The nine target contexts are processed in three consecutive rounds, with each context appearing once per round. This yields 27 adaptation stages, including nine initial visits and 18 revisits. Together, the three segments cover the original online adaptation split, with each frame entering the online stream only once. All online model states are maintained continuously throughout the stream without ground-truth context identities or transition boundaries. We evaluate all methods under one randomly sampled context order.

To quantify knowledge retention, we measure inter-visit forgetting as the detection performance degradation between consecutive visits to the same context. Let $E_{c,r}$ and $S_{c,r}$ denote the mean AP$_{\mathrm{3D}}$ of context $c$ at the end and beginning of visit $r$, respectively, averaged over IoU thresholds of 0.3 and 0.5. We compute the positive performance drop for each context before averaging.
Inter-visit forgetting is defined as
\begin{equation*}
\label{eq:sup_recurrence_forgetting}
F_r = \frac{1}{9}\sum_{c=1}^{9}\max\left(0, E_{c,r-1}-S_{c,r}\right), \quad r\in\{2,3\}.
\end{equation*}

As shown in Table~\ref{tab:sup_recurrence}, existing methods exhibit different limitations in balancing knowledge retention and continued adaptation. MOS~\cite{chen2025mos} and CTAOD~\cite{yoo2024and} achieve relatively strong final detection performance, with mean AP$_{\mathrm{3D}}$ of 56.25\% and 55.94\%, respectively, but exhibit substantial inter-visit forgetting. Specifically, their $F_2$/$F_3$ reach 7.50/11.42 pp and 11.61/17.52 pp, respectively. Similar degradation is observed for DPO~\cite{chen2024dpo} and CodeMerge~\cite{yang2025codemerge}, indicating that continued adaptation and historical model reuse do not necessarily preserve previously acquired detection capabilities. In contrast, BECoTTA~\cite{lee2024becotta} and Hybrid-TTA~\cite{park2025hybrid} exhibit relatively small forgetting, with $F_2$/$F_3$ of 0.23/0.34 pp and 0.67/2.02 pp, respectively, but achieve lower final detection accuracy of 44.47\% and 52.85\%. These results suggest that prioritizing knowledge retention may limit adaptation gains, whereas pursuing stronger adaptation through continuous updates may compromise previously acquired capabilities.

In contrast, KROMA achieves the highest mean AP$_{\mathrm{3D}}$ across all evaluated visit stages, with no observed inter-visit forgetting. Under the random context order, KROMA correctly reactivates the corresponding historical context in all revisits. This reliable reactivation is supported by compositional context recognition and semantic context memory, which enable KROMA to identify recurring conditions and recover their previously learned states. Upon reactivation, KROMA continues adapting the historical learner through context-specific residual updates, while knowledge-preserving mechanisms mitigate interference with previously acquired capabilities.
Importantly, mean AP$_{\mathrm{3D}}$ increases from 51.28\% after the first visit to 53.31\% and 58.25\% after the second and third visits, respectively. These improvements suggest that, as new observations arrive, reactivated historical knowledge provides a useful foundation for continued adaptation without limiting the model's ability to further adapt and acquire new knowledge.
Together, these results demonstrate the effectiveness of KROMA in combining context-aware knowledge reuse with knowledge-preserving online adaptation, enabling reliable knowledge reactivation and continued learning while maintaining high detection accuracy and stability across recurring driving conditions.

\subsection{Performance across Diverse Driving Conditions}
\label{sec:diverse_conditions}
\begin{figure}[t]
    \centering
    \includegraphics[
        width=0.5\textwidth
    ]{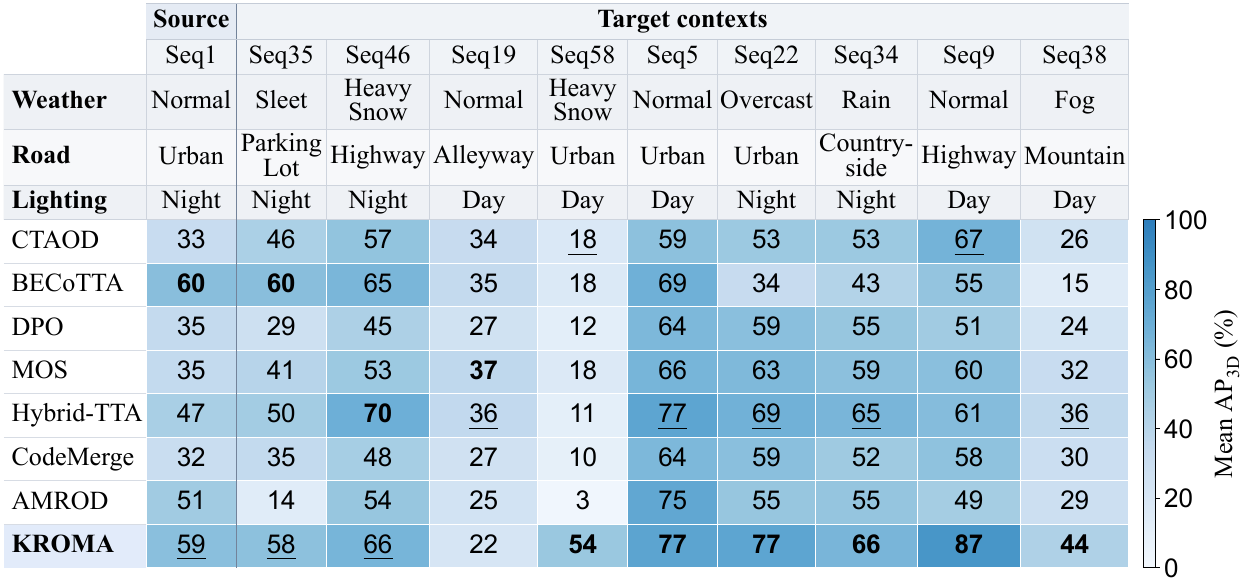}
\caption{\textbf{Per-context 3D detection performance.} Each cell reports the final mean $\mathrm{AP}_{\mathrm{3D}}$ (\%) after online adaptation, averaged over IoU thresholds of 0.3 and 0.5 and five randomly sampled context orders. Values are rounded to the nearest integer. The best and second-best results are highlighted in \textbf{bold} and \underline{underlined}.}
\label{fig:per_context_performance}
\vspace{-0.3em}
\end{figure}

Fig.~\ref{fig:per_context_performance} reports the final per-context detection performance in different driving conditions after sequential online adaptation, averaging $\mathrm{AP}_{\mathrm{3D}}^{0.3}$ and $\mathrm{AP}_{\mathrm{3D}}^{0.5}$.
KROMA ranks among the top two methods in nine of the ten evaluated contexts, with particularly pronounced advantages in several challenging environments.
Specifically, KROMA exhibits strong adaptability across challenging weather conditions and road types. It outperforms the strongest baselines by 35.50, 7.97, and 7.96~pp under heavy-snow (Seq58), foggy (Seq38), and overcast (Seq22) conditions, respectively.
On mountain roads, KROMA achieves a mean $\mathrm{AP}_{\mathrm{3D}}$ of 44.36\%, compared with 36.39\% for the strongest baseline, Hybrid-TTA~\cite{park2025hybrid}.
Beyond weather and road type, lighting variation presents another challenge for autonomous perception, particularly at night.
KROMA ranks among the top two methods in all five evaluated nighttime contexts, including a 7.96~pp improvement over the strongest baseline, Hybrid-TTA~\cite{park2025hybrid}, in overcast urban nighttime driving.
These results demonstrate that KROMA's competitive detection performance spans diverse combinations of weather, road type, and lighting, supporting its adaptability beyond any single environmental setting.

\subsection{Ablation Studies}
\label{sec:ablation}
To evaluate the contributions of KROMA's key components, we compare the full framework with six variants under a common experimental protocol, as shown in Table~\ref{tab:ablation}.
We first examine three broader configurations to assess the overall contributions of the two main modules.
(1) \emph{w/o Both Modules} removes both modules and sequentially updates a single detector across contexts.
(2) \emph{w/o Compositional Context} replaces compositional semantic context recognition with a KMeans-based context classifier trained on the same data.
(3) \emph{w/o Knowledge Preservation} uses context recognition to independently train and store a model for each context, without applying the proposed knowledge-preserving update strategy.
We then examine three key mechanisms of online adaptation: historical state retrieval, parameter superposition, and knowledge preservation.
Specifically, (4) \emph{w/o Historical Retrieval} inherits the previously active context's retained state instead of selecting among historical candidates.
(5) \emph{w/o Parameter Superposition} replaces superposed weight updates in the fusion network with conventional adapter-based updates.
(6) \emph{w/o Preservation Mechanisms} removes the knowledge-preservation mechanisms, including knowledge-constrained learning and selective state retention.

\begin{table}[t]
\centering
\caption{
    \textbf{Ablation study.}
    Mean AP$_{\mathrm{BEV}}$ and Mean AP$_{\mathrm{3D}}$ across ten contexts.
    Results are mean $\pm$ std over five context orders.
}
\label{tab:ablation}

\footnotesize
\setlength{\tabcolsep}{1.5pt}
\renewcommand{\arraystretch}{1.0}

\setbox0=\hbox{\textbf{00}}
\edef\AblDigitsWidth{\the\wd0}
\setbox0=\hbox{\textbf{.}}
\edef\AblDotWidth{\the\wd0}

\newcommand{\AblNum}[2]{%
    \makebox[\AblDigitsWidth][r]{#1}%
    \makebox[\AblDotWidth][c]{.}%
    \makebox[\AblDigitsWidth][l]{#2}%
}

\begin{tabular}{@{}
    >{\raggedright\arraybackslash}p{0.44\columnwidth}
    @{\hspace{3pt}}
    r@{\,}c@{\,}l
    @{\hspace{5pt}}
    r@{\,}c@{\,}l
    @{}
}
\toprule
Method
& \multicolumn{3}{c}{Mean AP$_{\mathrm{BEV}} $ }
& \multicolumn{3}{c}{Mean AP$_{\mathrm{3D}}$} \\
\midrule

w/o Both Modules
& \AblNum{58}{51} & $\pm$ & \AblNum{7}{51}
& \AblNum{34}{06} & $\pm$ & \AblNum{10}{67} \\

\addlinespace[0pt]

w/o Compositional Context
& \AblNum{63}{59} & $\pm$ & \AblNum{2}{57}
& \AblNum{41}{17} & $\pm$ & \AblNum{3}{56} \\

\addlinespace[0pt]

w/o Knowledge Preservation
& \AblNum{65}{89} & $\pm$ & \AblNum{0}{00}
& \AblNum{57}{74} & $\pm$ & \AblNum{0}{01} \\

\midrule

w/o Historical Retrieval
& \AblNum{70}{57} & $\pm$ & \AblNum{2}{16}
& \AblNum{60}{75} & $\pm$ & \AblNum{4}{48} \\

\addlinespace[0pt]

w/o Parameter Superposition
& \AblNum{67}{62} & $\pm$ & \AblNum{0}{34}
& \AblNum{59}{48} & $\pm$ & \AblNum{0}{56} \\

\addlinespace[0pt]

w/o Preservation Mechanisms
& \AblNum{66}{26} & $\pm$ & \AblNum{0}{76}
& \AblNum{57}{42} & $\pm$ & \AblNum{0}{40} \\

\midrule

KROMA (Full)
& \textbf{\AblNum{70}{59}} & $\pm$ & \AblNum{1}{03}
& \textbf{\AblNum{61}{09}} & $\pm$ & \AblNum{0}{95} \\
\bottomrule
\end{tabular}
\end{table}

Full KROMA achieves the highest mean $\mathrm{AP}_{\mathrm{BEV}}$ and $\mathrm{AP}_{\mathrm{3D}}$ of 70.59\% and 61.09\%, respectively.
Removing both modules reduces these scores to 58.51\% and 34.06\%, demonstrating their substantial combined benefits.
Replacing compositional context recognition with clustering reduces BEV and 3D AP by 7.00 and 19.92~pp, respectively, supporting the importance of semantic context identification.
Independent Context Models (w/o Knowledge Preservation) perform 4.70 and 3.35~pp below KROMA, supporting the effectiveness of knowledge reuse and preservation.
Among the individual mechanisms, historical retrieval primarily improves adaptation stability.
Inheriting the previously active context's state increases the cross-order standard deviation of mean $\mathrm{AP}_{\mathrm{3D}}$ from 0.95 to 4.48~pp, as consecutive driving conditions may differ substantially.
This indicates that selecting the most suitable historical state before adaptation can indeed reduce this instability.
Replacing parameter superposition with conventional adapters lowers mean $\mathrm{AP}_{\mathrm{3D}}$ by 1.61~pp, demonstrating the advantage of parameter superposition.
Removing knowledge preservation mechanisms reduces mean BEV and 3D AP by 4.33 and 3.67~pp, respectively, highlighting the importance of constraining updates and retaining acquired knowledge.
Together, these findings further demonstrate the significance and specific contributions of KROMA's key components.

\subsection{Hyperparameter Sensitivity Analysis}
\label{sec:sup_hyperparameters}

To evaluate the sensitivity of KROMA to hyperparameter choices, we examine three parameters: the confidence threshold $\tau$ and temporal consistency window $W$ for context recognition, and the validation memory capacity $M_v$ for selective state retention. Starting from the default configuration ($\tau=0.9$, $W=5$, and $M_v=32$), we vary each parameter independently while keeping the remaining parameters and experimental settings unchanged. All experiments are conducted under the same randomly sampled context order, following the same online adaptation and evaluation protocols as the main experiments. We report detection performance macro-averaged across the nine target contexts in Table~\ref{tab:sup_hyperparameters}.

\begin{table}[t]
\centering
\caption{
    \textbf{Hyperparameter sensitivity analysis.}
    Detection performance under a single randomly sampled context order, macro-averaged across nine target contexts.
    The default settings are $\tau=0.9$, $W=5$, and $M_v=32$.
    Each parameter is varied independently while the others remain at their default values.
}
\label{tab:sup_hyperparameters}

\footnotesize
\setlength{\tabcolsep}{3pt}
\renewcommand{\arraystretch}{0.9}

\begin{tabular}{@{}llcccc@{}}
\toprule
\multirow{2}{*}{Parameter} &
\multirow{2}{*}{Value} &
\multicolumn{2}{c}{AP$_{\mathrm{BEV}}$ (\%) $\uparrow$} &
\multicolumn{2}{c}{AP$_{\mathrm{3D}}$ (\%) $\uparrow$} \\
\cmidrule(lr){3-4}
\cmidrule(lr){5-6}
& & IoU 0.3 & IoU 0.5 & IoU 0.3 & IoU 0.5 \\
\midrule

& 0.80 & 70.90 & 60.03 & 69.39 & 45.92 \\
\rowcolor{black!3}
$\tau$ & 0.90 & 76.51 & 66.74 & 75.38 & 48.51 \\
& 0.95 & 73.14 & 63.59 & 71.48 & 44.08 \\

\midrule

& 3 & 76.51 & 66.74 & 75.38 & 48.51 \\
\rowcolor{black!3}
$W$ & 5 & 76.51 & 66.74 & 75.38 & 48.51 \\
& 7 & 76.51 & 66.74 & 75.38 & 48.51 \\

\midrule

& 16 & 73.27 & 62.62 & 70.48 & 44.92 \\
\rowcolor{black!3}
$M_v$ & 32 & 76.51 & 66.74 & 75.38 & 48.51 \\
& 64 & 76.33 & 67.32 & 74.64 & 51.20 \\

\bottomrule
\end{tabular}
\end{table}

As shown in Table~\ref{tab:sup_hyperparameters}, KROMA maintains relatively stable detection performance across the evaluated hyperparameter settings, with mean AP$_{\mathrm{3D}}$ ranging from 57.66\% to 62.92\%. For the confidence threshold $\tau$, the default value of 0.9 achieves the best performance across all four AP metrics, reaching a mean AP$_{\mathrm{3D}}$ of 61.95\%, while both lower and higher thresholds maintain a mean AP$_{\mathrm{3D}}$ close to 58\%.
The superior performance at $\tau=0.9$ may be attributed to a better balance between filtering out unreliable context predictions and enabling timely context recognition and knowledge activation. 
Meanwhile, varying the temporal consistency window $W$ from 3 to 7 yields identical results across all four AP metrics when rounded to two decimal places. These results indicate that KROMA maintains stable detection performance across different temporal window sizes.

For the validation memory capacity $M_v$, increasing the capacity from 16 to 32 and 64 improves mean AP$_{\mathrm{3D}}$ from 57.70\% to 61.95\% and 62.92\%, respectively. These results suggest that incorporating more observations into candidate state evaluation can benefit online adaptation, but the performance gain becomes smaller as the capacity increases. The default capacity of 32 offers a favorable balance between detection performance and memory requirements. Overall, the sensitivity analysis demonstrates that KROMA maintains relatively stable detection performance across the evaluated hyperparameter settings, while appropriate choices of the confidence threshold and validation memory capacity can further improve online adaptation.

\begin{figure}[h]
    \centering
    \includegraphics[
        width=0.75\linewidth
    ]{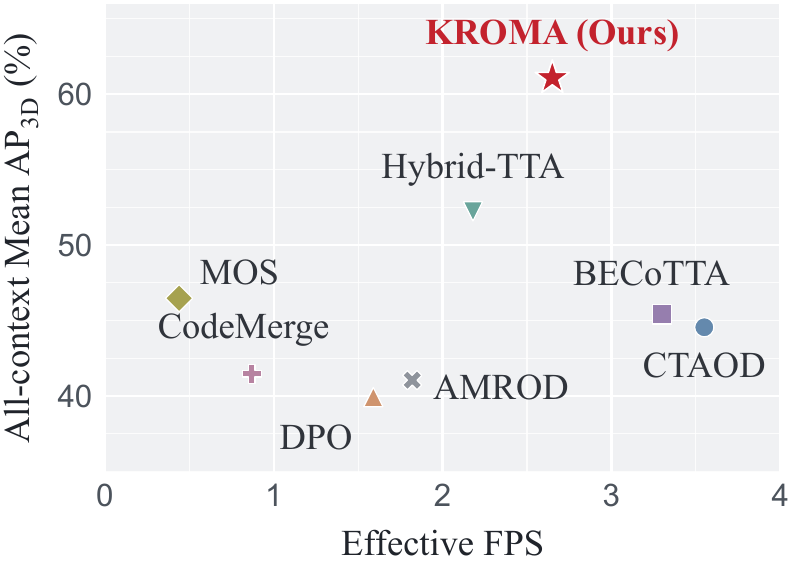}
    \caption{
    \textbf{Accuracy--Efficiency Trade-off.}
    Mean $\mathrm{AP}_{\mathrm{3D}}$ is averaged across ten contexts.
    Effective FPS includes online adaptation and target inference.
    Results are averaged over five context orders.
}
    \label{fig:accuracy_efficiency}
\end{figure}

\subsection{Efficiency Analysis}
\label{sec:efficiency}

To assess the computational cost of KROMA's detection gains, we jointly evaluate final detection accuracy and effective throughput in Fig.~\ref{fig:accuracy_efficiency}.
Effective FPS divides the number of target inference frames by the combined time for online adaptation and target inference, including context recognition, historical retrieval, parameter updates, and model-state management.
KROMA achieves the highest all-context mean $\mathrm{AP}_{\mathrm{3D}}$ of 61.09\% at 2.65 effective FPS, attaining the third-highest throughput among the eight methods.
Compared with Hybrid-TTA~\cite{park2025hybrid}, the strongest baseline in detection accuracy, KROMA improves mean $\mathrm{AP}_{\mathrm{3D}}$ by 8.85~pp while increasing effective throughput by 21.6\%, from 2.18 to 2.65 FPS.
Although CTAOD~\cite{yoo2024and} and BECoTTA~\cite{lee2024becotta} achieve higher throughput at 3.55 and 3.30 FPS, their mean $\mathrm{AP}_{\mathrm{3D}}$ scores are lower than KROMA's by 16.53 and 15.67~pp, respectively.
These results demonstrate a favorable accuracy--efficiency trade-off: KROMA delivers the highest detection accuracy while maintaining competitive throughput, even when accounting for the computational overhead of knowledge retrieval and preservation.


\section{Conclusion}
\label{sec:conclusion}

In this paper, we presented KROMA, an online adaptation method for multimodal 3D object detection under changing driving conditions.
KROMA enables context-aware knowledge reuse through compositional semantic context recognition and detection-guided historical state retrieval.
Building on retrieved knowledge, context-specific residual adaptation and knowledge constraints mitigate interference and preserve previously acquired capabilities, while selective state retention maintains reusable model states for future adaptation.
Experiments across ten K-Radar contexts demonstrate that KROMA achieves
$\mathrm{AP}_{\mathrm{3D}}^{0.3}$ and $\mathrm{AP}_{\mathrm{3D}}^{0.5}$
of 73.55\% and 48.62\%, outperforming the strongest baseline at each
threshold by 7.7 and 9.9 percentage points, respectively.
These improvements are accompanied by greater adaptation stability
and competitive computational efficiency.
These results highlight the importance of combining relevant historical knowledge reuse with knowledge-preserving adaptation for reliable multimodal perception under evolving driving conditions.

\section*{Acknowledgements}
This work has been funded in part by NSF, with award numbers \#2112665, \#2112167, \#2003279, \#2120019, \#2211386, \#2052809, \#1911095 and in part by PRISM and CoCoSys, centers in JUMP 2.0, an SRC program sponsored by DARPA.


{
    \small
    \bibliographystyle{ieeenat_fullname}
    \bibliography{ref}
}


\end{document}